# From Prototype to Massing

**Yiliang Shao**
Skidmore, Owings & Merrill

## Learning Graph Correspondence for Scalable Facade Modeling

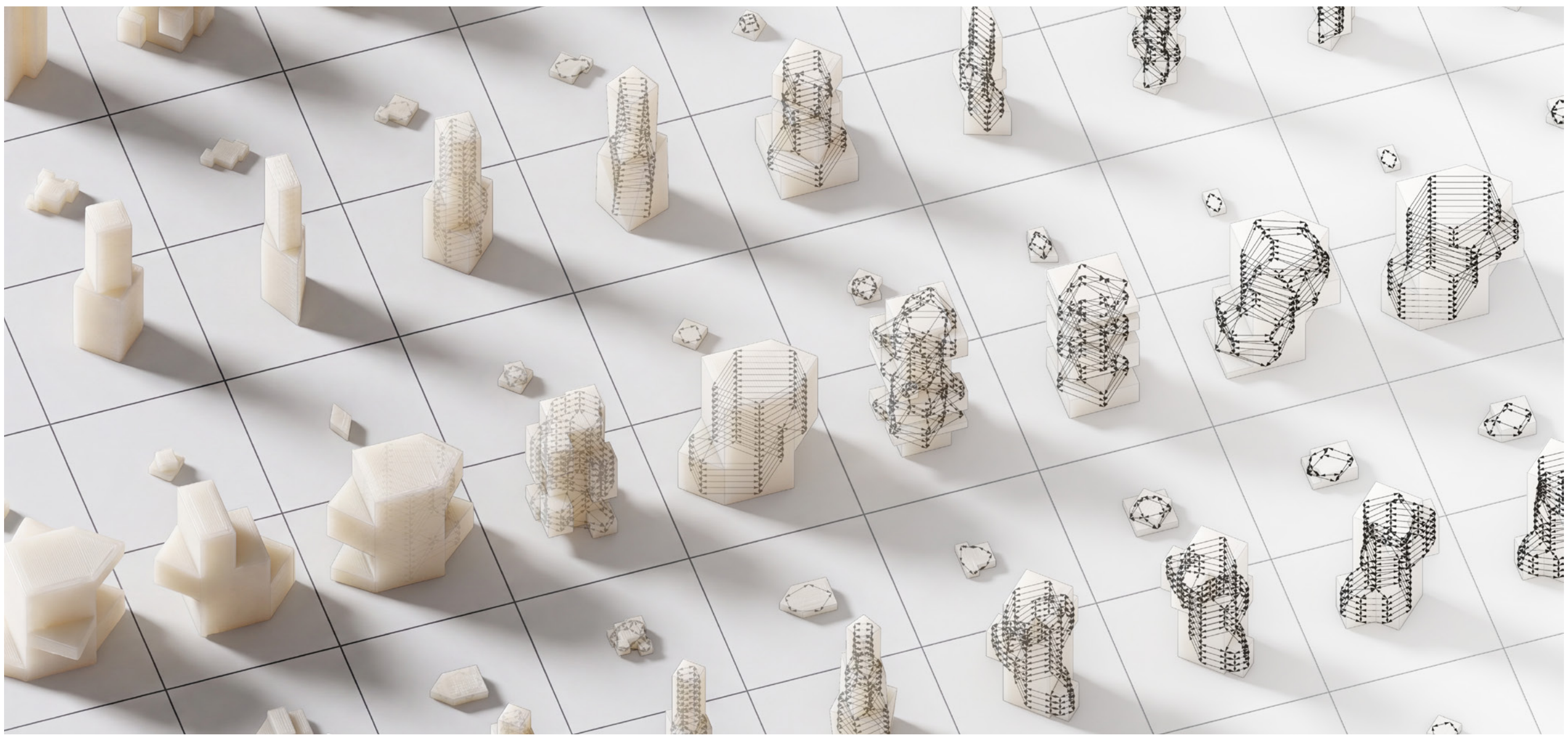

1

### ABSTRACT

Facade articulation at building scale is usually resolved manually or through parametric systems that encode design intent as explicit rules. This paper investigates a complementary workflow in which a designer-authored prototype supplies transferable articulation logic. Prototype and target massing are represented as heterogeneous attributed graphs, and a Siamese graph attention network with instance-local triplet supervision embeds their segments in a shared metric space. Many-to-one nearest-neighbor retrieval then routes prototype Patterns to the larger target. On the original 85/15 within-corpus holdout, prototype-to-massing top-1 accuracy reached 96.5%, compared with a reported 76.5% raw-feature baseline. A Rhino demonstrator converts predicted correspondence into reusable facade Patterns, and qualitative applications on four built projects show the workflow across distinct massing organizations. The contribution is not automated facade generation, but an authoring model in which resolved geometry serves as a computational specification for project-scale propagation.

Keywords: facade articulation transfer, example-based modeling, heterogeneous graph attention network, triplet metric learning, graph correspondence

## INTRODUCTION

Two approaches dominate facade articulation in practice: manual modeling, and project-specific parametric rules. Manual replication is costly at building scale. Parametric authoring provides precise control when variation can be anticipated, but requires designers to formalize that variation in advance (Kolarevic 2003; Woodbury 2010); visual programming systems such as Grasshopper and Dynamo are its current instruments. Intent is often resolved first in a representative bay or elevation fragment, and reusing that fragment across a larger envelope still requires replication or reconstruction as a rule system. Between resolved local geometry and project-scale application there remains a transfer problem.

This paper investigates an exemplar-driven alternative in which the designer-authored prototype supplies the transferable logic. The method learns correspondence between relational abstractions of the prototype and of a larger target massing; predicted matches then route detailed components onto the target without a global semantic taxonomy or a project-wide propagation script. It complements parametric authoring rather than replacing it.

Both inputs become heterogeneous attributed graphs, whose nodes represent geometric segments and whose horizontal and vertical adjacencies form typed edge sets. A shared graph attention encoder maps both graphs into a metric space where pair-local counterparts approach one another, and nearest-neighbor retrieval allows prototype identities to repeat across the larger target. The graph stores scaled geometry descriptors and topology, not facade-module geometry or world coordinates; detailed content remains designer-authored and is routed downstream (Figure 2).

The contributions of this research are:

- a prototype-driven facade propagation workflow in which a single designer-authored exemplar drives transfer of its detailing logic across the full massing envelope.
- a heterogeneous graph abstraction that converts building massing geometry into typed relational structures, encoding sequential facade rhythm and vertical stacking as separate edge types.
- and an instance-local supervision formulation that learns node-level correspondence relative to each prototype-massing pair rather than to a dataset-wide label vocabulary or a predefined architectural taxonomy, realized here with a Siamese heterogeneous graph attention encoder.

On the original within-corpus holdout, the complete model reached 96.5% prototype-to-massing top-1 accuracy, about 20 points above direct retrieval on the same descriptors. Together with the Rhino demonstrator and built-project cases, this supports a technically viable workflow in which resolved geometry supplies the correspondence vocabulary for a compatible envelope. The evidence does not extend to production reliability or to generalization beyond the authored corpus.

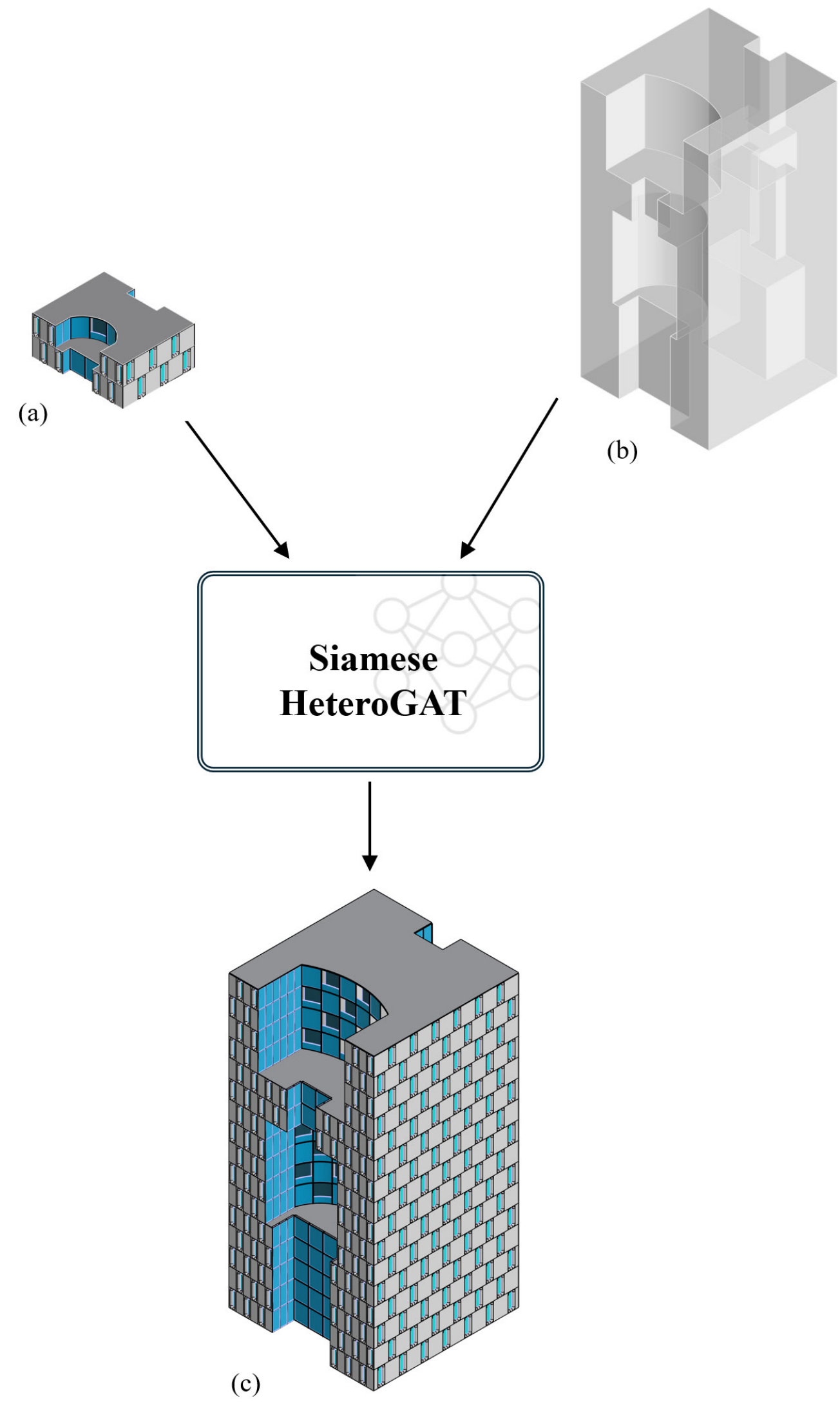

2

2 System overview: (a) Localized prototype. (b) Target massing. (c) Propagated facade via learned correspondence.

## RELATED WORK

### Computational Design and Procedural Facade Systems

Architectural computation has long pursued methods for encoding facade articulation as reusable generative logic (Kolarevic 2003; Woodbury 2010). Visual programming environments and procedural modeling pipelines build on shape grammar formalisms (Stiny and Gips 1972), split grammars (Wonka et al. 2003), and procedural facade systems such as CGA Shape (Müller et al. 2006; Schwarz and Müller 2015), and they allow designers to construct generative systems through parameter graphs (Aish and Woodbury 2005) and symbolic production rules. Within their operating envelope these systems are powerful: when the space of intended variation can be anticipated and formalized, deterministic rule structures provide precise and maintainable control.

They share one limitation: design logic must be structured explicitly and in advance. Articulation that was resolved locally and empirically, in a bay study or a representative elevation, cannot be reused directly; it must first

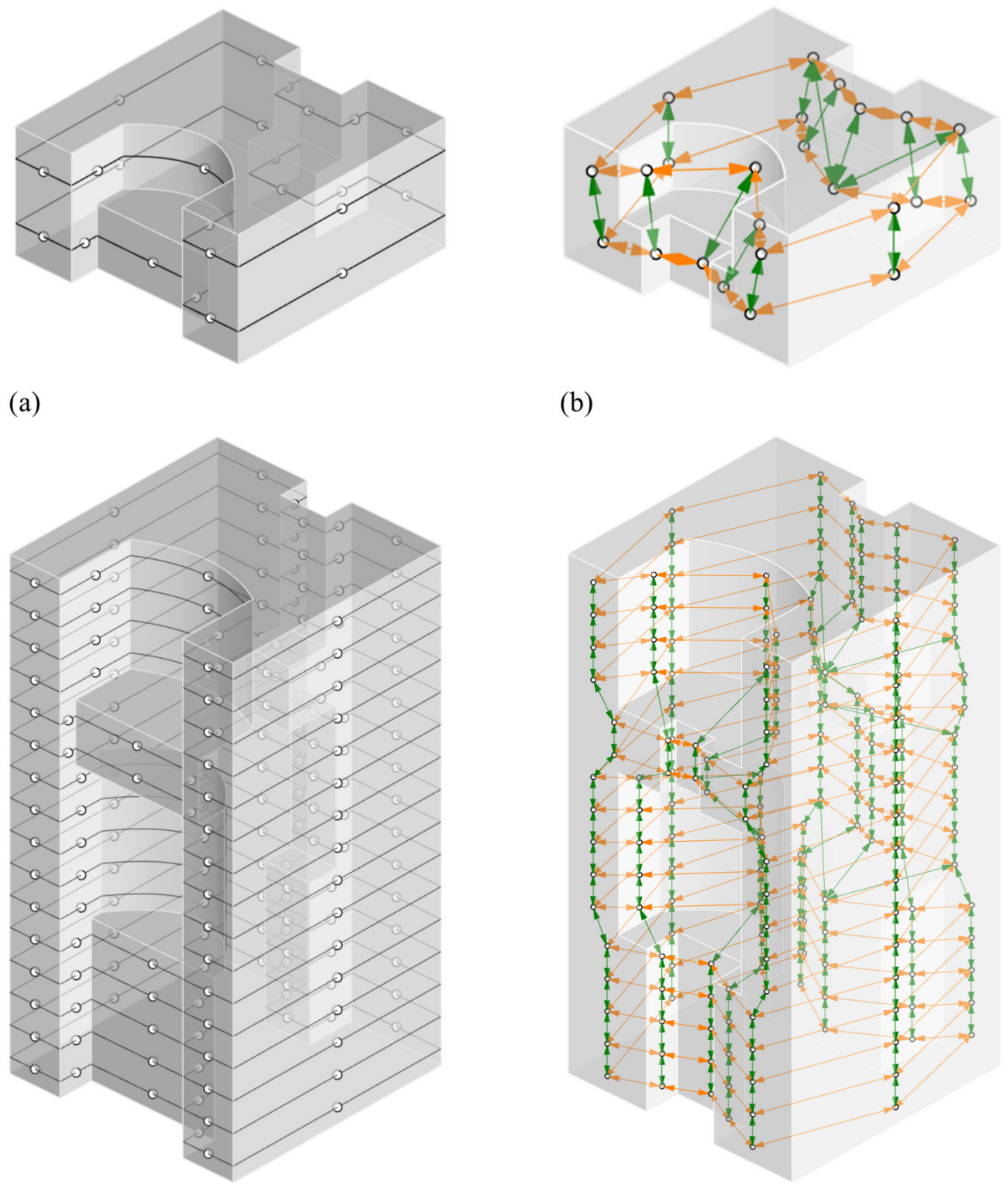

(a) (b) (c) (d)

3

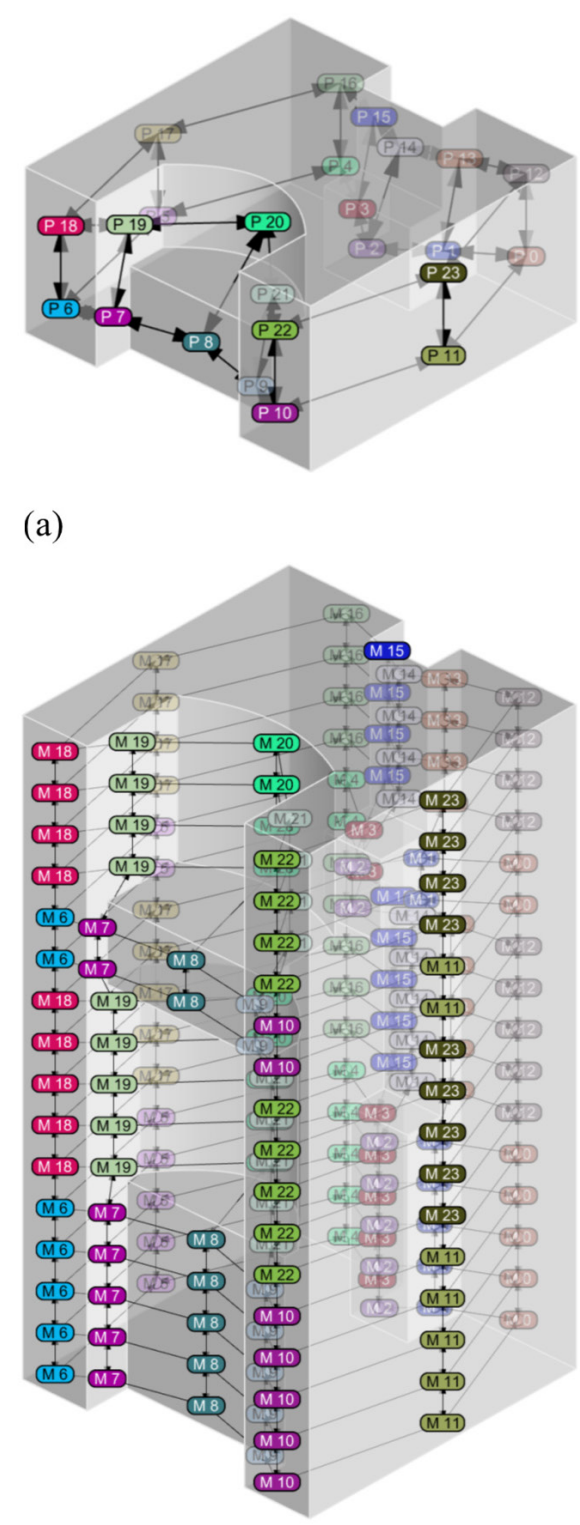

(a) (b)

4

3 Heterogeneous graph construction. (a) Prototype envelope segmented into contour-level segments. (b) Prototype graph with horizontal and vertical edges. (c) Massing envelope segmented under the same rule. (d) Massing graph, illustrating the cardinality asymmetry $|V_S| \ll |V_Q|$.

4 Instance-local labels. (a) Prototype graph with unique correspondence identifiers. (b) Massing graph: target segments inherit prototype labels, with reuse across multiple segments enabling many-to-one transfer.

be reconstructed as a rule set. This reconstruction step is the specific cost the present work addresses. Example-based methods in graphics avoid it by generalizing from a supplied exemplar pair rather than from an explicit rule (Hertzmann et al. 2001).

### Graph Representations and Learning-Based Correspondence

Graph representations in architectural computation have been primarily analytical, applied to a single model for space syntax analysis, floorplan generation, point-cloud reconstruction, or buildability assessment (Hillier and Hanson 1984; Hu et al. 2020; Chen et al. 2024; Wang et al. 2026). In the present work the graph instead serves as a transfer medium between abstractions of unequal scale, with correspondence itself as the output, a use not represented in the graph-learning work reviewed above.

Outside architectural computation, relational correspondence has been reformulated as a metric learning problem. Siamese architectures (Bromley et al. 1993) trained with triplet objectives (Schroff, Kalenichenko, and Philbin 2015) learn shared embedding spaces where corresponding elements cluster and non-matching elements separate, enabling flexible correspondence between unequal graphs through nearest-neighbor retrieval. Graph Matching Networks (Li et al. 2019) develop this paradigm specifically for graph-structured objects, computing pairwise similarity through cross-graph attention over jointly reasoned graph pairs.

Where graph-based learning has been applied to architectural problems, supervision has consistently been organized around globally stable label vocabularies such as room types or module categories, fixed across the entire dataset (Wu et al. 2019; Hu et al. 2020; Nauata et al. 2020). Facade articulation transfer is not that task: the massing is matched against a specific designer-authored prototype, not a universal vocabulary. The correspondence problem is pair-local, and requires a supervision structure defined relative to each prototype instance.

## REPRESENTING BUILDING ENVELOPES AS ASYMMETRIC GRAPH PAIRS

We frame prototype-driven facade propagation as a pair-local correspondence task: the system learns how patterns in a designer-authored prototype map onto structurally analogous segments in a larger massing.

### Asymmetric Graph Formulation

To compute this mapping, both geometric domains are abstracted into heterogeneous attributed graphs. Each design task is formulated as a paired representation $(G_S, G_Q)$, where $G_S$ is the support graph derived from the smaller prototype model, and $G_Q$ is the query graph derived from the larger target massing model. Each graph is defined as:

$$G = (V, X, \{E_\phi\}, \{X_{e,\phi}\}, Y)$$

where $V$ represents discrete geometric segments, $X \in \mathbb{R}^{|V| \times 7}$ encodes intrinsic node features, $E_\phi$ and $X_{e,\phi}$ denote typed relational edge sets and their attributes, and $Y$ stores correspondence supervision targets.

The two graphs differ substantially in scale and cardinality, though both are assumed to share the same level of representational detail: the prototype, smaller in extent, must still cover every visible design element present on the massing (Figure 3):

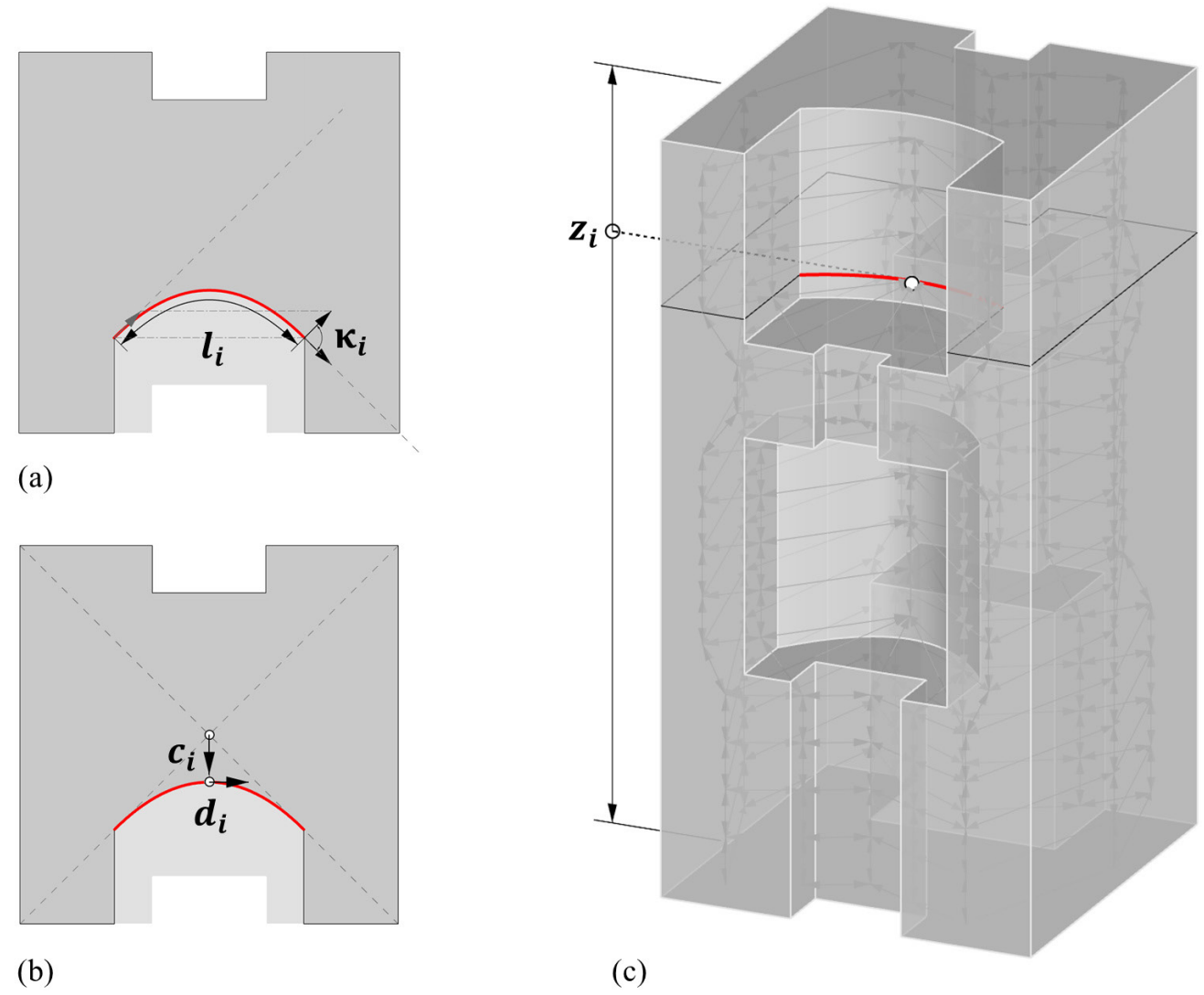

5 Node feature components. (a) Intrinsic geometry: normalized relative length $l_i$ and tangent deviation $\kappa_i$. (b) Absolute orientation $d_i$ and relative positioning $c_i$. (c) Normalized elevation $z_i$ within the building frame. 5

$$|V_S| \ll |V_Q|$$

Classical graph isomorphism seeks a bijection, while subgraph isomorphism seeks an injective match from a smaller graph into part of a larger one. Neither directly represents the deployment mapping here, in which multiple massing nodes may reuse the same prototype identity. We therefore learn correspondence relative to each prototype-massing pair without assuming a fixed facade-element taxonomy.

### Instance-Local Labels

Since the massing is matched against a specific designer-authored prototype rather than a universal architectural vocabulary, the meaning of each label is local to one graph pair. The same integer may identify unrelated regions in another pair, so no fixed dataset-wide class vocabulary exists, and a global classifier would conflate instance-specific identities. We use metric supervision defined within each pair instead: equal labels identify positive prototype-massing correspondences, unequal labels identify negatives, and shared encoder weights learn how those pair-local roles are expressed through node geometry and neighborhood structure (Figure 4). This formulation avoids requiring a global taxonomy. It does not establish that no instance-conditioned classifier could work.

The label vocabulary remains deliberately fine-grained: nearly identical prototype regions may receive distinct identifiers rather than a shared class, even when they could later route to the same facade Pattern. This encourages the encoder to preserve pair-local distinctions and to attend to relational position instead of collapsing all locally similar segments. Pattern assignments, however, were authored downstream in Rhino and are not stored in the graph corpus. The evaluation reported here therefore cannot determine how often a label-level confusion preserves the instantiated Pattern and how often it changes it; that consequence is treated as a limitation rather than assumed to be harmless.

### Node Features and Edge Attributes

Nodes are discrete geometric segments extracted by slicing the input B-Rep at regular vertical intervals and subdividing at geometric inflection points. For each segment, a 7-dimensional descriptor vector is computed, normalized to $[0,1]$. The descriptor encodes intrinsic geometry (normalized relative length $l_i$ and tangent deviation $\kappa_i$), absolute orientation ($d_i$, two scaled planar components), and relative positioning ($c_i$, a 2-dimensional centroid-relative vector, and normalized elevation $z_i$, locating each segment within the building's organizational frame) (Figure 5).

Edges are instantiated as two types, reflecting the different behavior along sequential and stacking axes. Horizontal edges ($E_H$) connect segments sequentially along a single contour level, each carrying a 3-dimensional feature vector encoding traversal direction, angular transition, and relative scale. Vertical edges ($E_V$) encode stacking logic between successive slicing levels, each carrying a 4-dimensional feature vector encoding vertical directionality, sectional depth offset, centroid alignment, and length consistency. Raw XYZ coordinates are not stored. Orientation and relative-position features retain the preprocessing frame, so the representation reduces dependence on translation and scale but is not rotation-invariant.

## SIAMESE HETEROGAT WITH TRIPLET SUPERVISION

### Siamese Architecture

A shared encoder processes both graphs, organizing one metric space by correspondence rather than graph domain (Figure 6). A linear projection maps seven node features to the hidden space, followed by two heterogeneous attention layers with residual paths. A final projection produces an L2-normalized embedding, so Euclidean retrieval depends on angular separation rather than vector magnitude. The two layers broaden the receptive field beyond immediate neighbors; depth was not isolated experimentally.

### Heterogeneous Graph Attention Layer

Each layer separates horizontal and vertical message passing through relation-specific transformations (Schlichtkrull et al. 2018), with separate GAT parameters and a linear edge encoder in each branch, projecting 3-D horizontal and 4-D vertical edge attributes. Edge-conditioned attention is computed independently per relation, combining message passing (Gilmer et al. 2017) with attention aggregation (Veličković et al. 2018); projected edge and node features enter the attention score so each axis can weight geometric transitions. The streams are then summed with a projected residual, supplying an identity-like path, and passed through ReLU.

### Triplet Metric Learning

Because labels carry no shared semantics across graph pairs, there is no fixed dataset-wide classification target. We therefore optimize a Triplet Margin Loss (Schroff, Kalenichenko, and Philbin 2015). For a prototype anchor, a positive is a massing node with the same pair-local label, while a negative carries a different label within that massing.

$$\mathscr{L}_{\text{triplet}} = \max\left(0, \|f_\theta(x_a) - f_\theta(x_p)\|_2 - \|f_\theta(x_a) - f_\theta(x_n)\|_2 + \alpha\right)$$

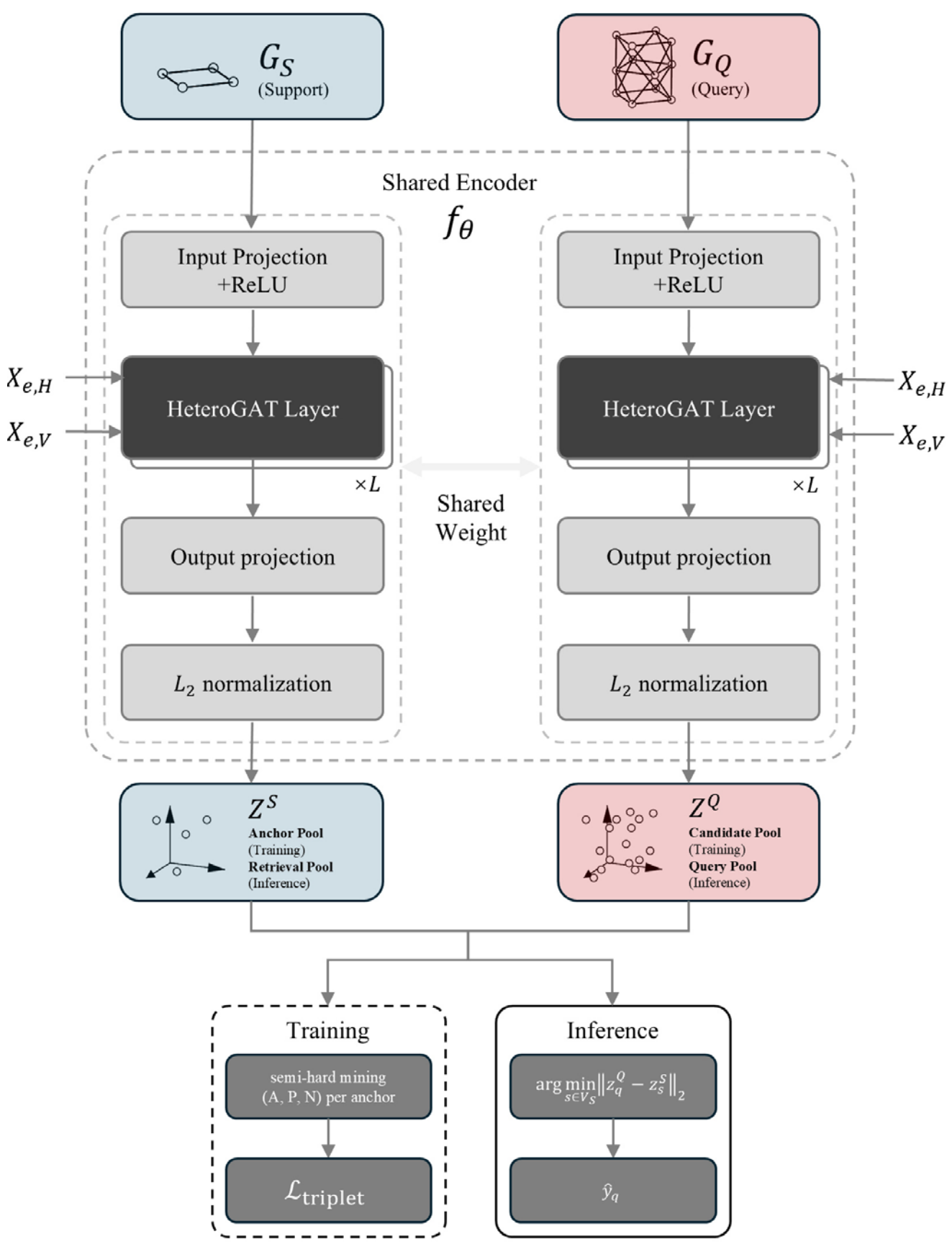

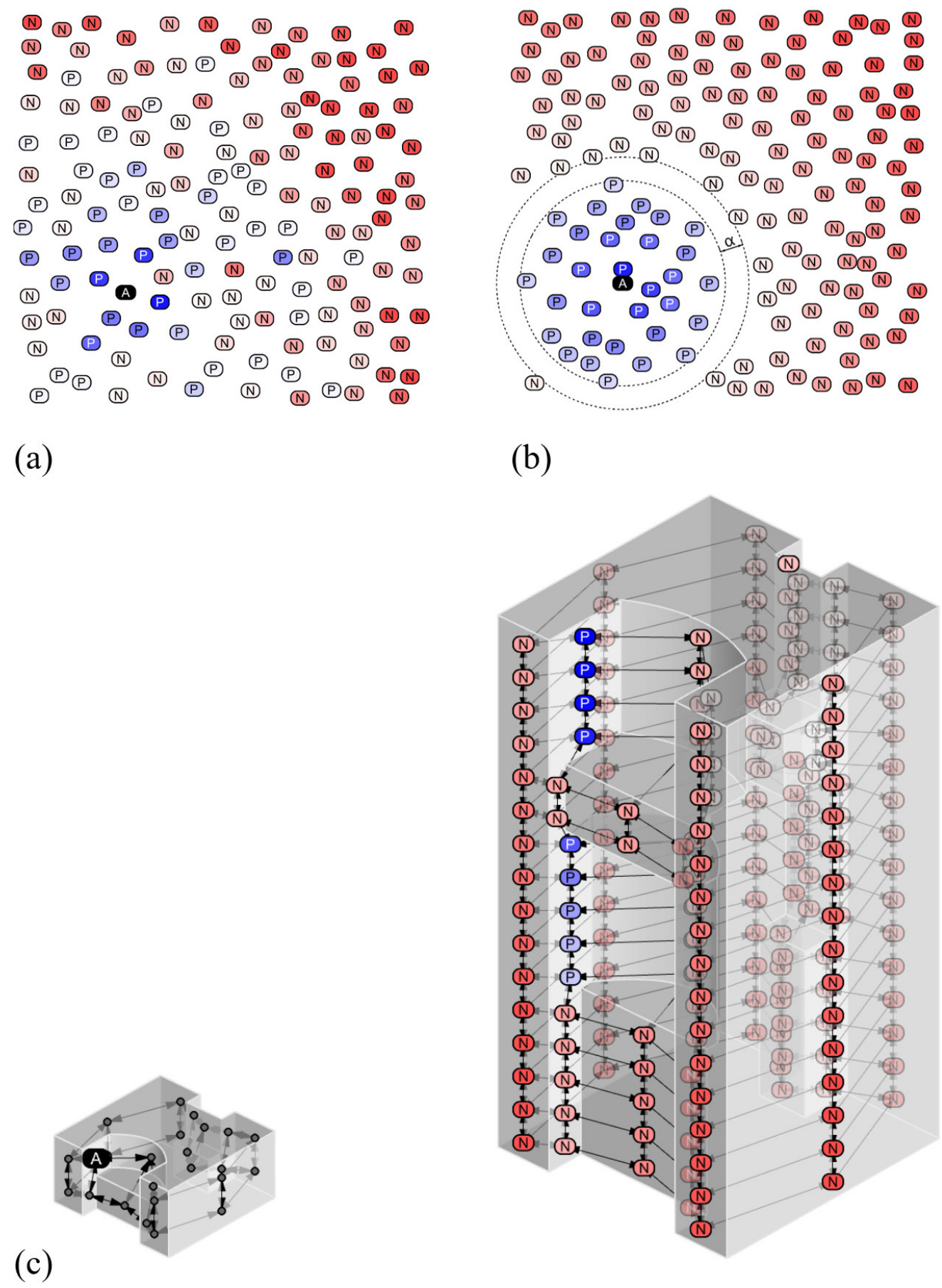

6

6 Siamese network architecture. A shared encoder $f_\theta$ embeds support graph $G_S$ and query graph $G_Q$ into a shared embedding space via heterogeneous graph attention layers conditioned on typed horizontal ($X_{e,H}$) and vertical ($X_{e,V}$) edge attributes. Training optimizes a triplet loss under staged random, semi-hard, and hard negative mining; inference assigns each query node the label of its nearest support neighbor in the embedding space.

7

7 Triplet sampling and metric structure. (a) Warmup phase: random negative selection, positives and negatives intermixed in the embedding space around the anchor (A). (b) Semi-hard mining phase: positives have clustered within the margin under triplet supervision; negatives are mined from the boundary region, farther than the positive but within the margin of it (dashed); negatives already beyond the margin contribute nothing to the loss. (c) The same anchor–positive–negative roles mapped onto a support–query graph pair.

Negatives partition into three regions defined relative to the anchor–positive distance $d_{pos}$ and margin $\alpha$: hard negatives lie closer to the anchor than the positive, semi-hard negatives sit between $d_{pos}$ and $d_{pos} + \alpha$, and easy negatives lie beyond the margin and produce zero loss.

For each valid prototype anchor, positives are sampled uniformly from same-label massing nodes; all different-label nodes form the negative pool. Epochs 1-20 sample uniformly. In epochs 21-120, semi-hard candidates satisfy $d_{pos} < d_{neg} < d_{pos} + 0.5$; one is drawn uniformly from the closest k, with k decreasing from 40 to 8, or the nearest negative if none qualifies. Epochs 121-150 draw uniformly from the closest k negatives as k decreases from 10 to 5; later epochs keep k=5 (Figure 7).

Models were implemented in PyTorch 2.10 and PyTorch Geometric 2.7 (Paszke et al. 2019; Fey and Lenssen 2019). Training used two heterogeneous attention layers with separate horizontal and vertical four-head GATConv branches. Each head has 512 channels; branch outputs and the residual are 2,048-D, followed by a 256-D L2-normalized embedding. Edge attributes project to 512-D, self-loops are disabled, and dropout is 0.1. Euclidean triplet loss used margin 0.5. Adam used learning rate $2 \times 10^{-3}$, weight decay $10^{-5}$, gradient clipping 1.0, one pair per step, and eight-pair accumulation. ReduceLROnPlateau, monitoring training loss, halved the rate after 20 stagnant epochs to a floor of $10^{-5}$. Validation selection used a 0.001 improvement threshold, grace through epoch 150, patience 30, and a 200-epoch cap. The original split indices and RNG state did not survive, so that run cannot be exactly re-executed.

### Inference: Nearest-Neighbor Label Transfer

At inference, each query node takes the label of its nearest node in the other graph. Retrieval is local, so prototype labels may be reused across a larger massing. The historical accuracy measure queries prototype nodes against the massing; deployment reverses the query, so each massing segment retrieves a prototype Pattern. Hard one-to-one assignment is incompatible with repeated reuse, although soft transport remains possible (Kuhn 1955; Cuturi 2013).

## EVALUATING CORRESPONDENCE ACCURACY

### Dataset

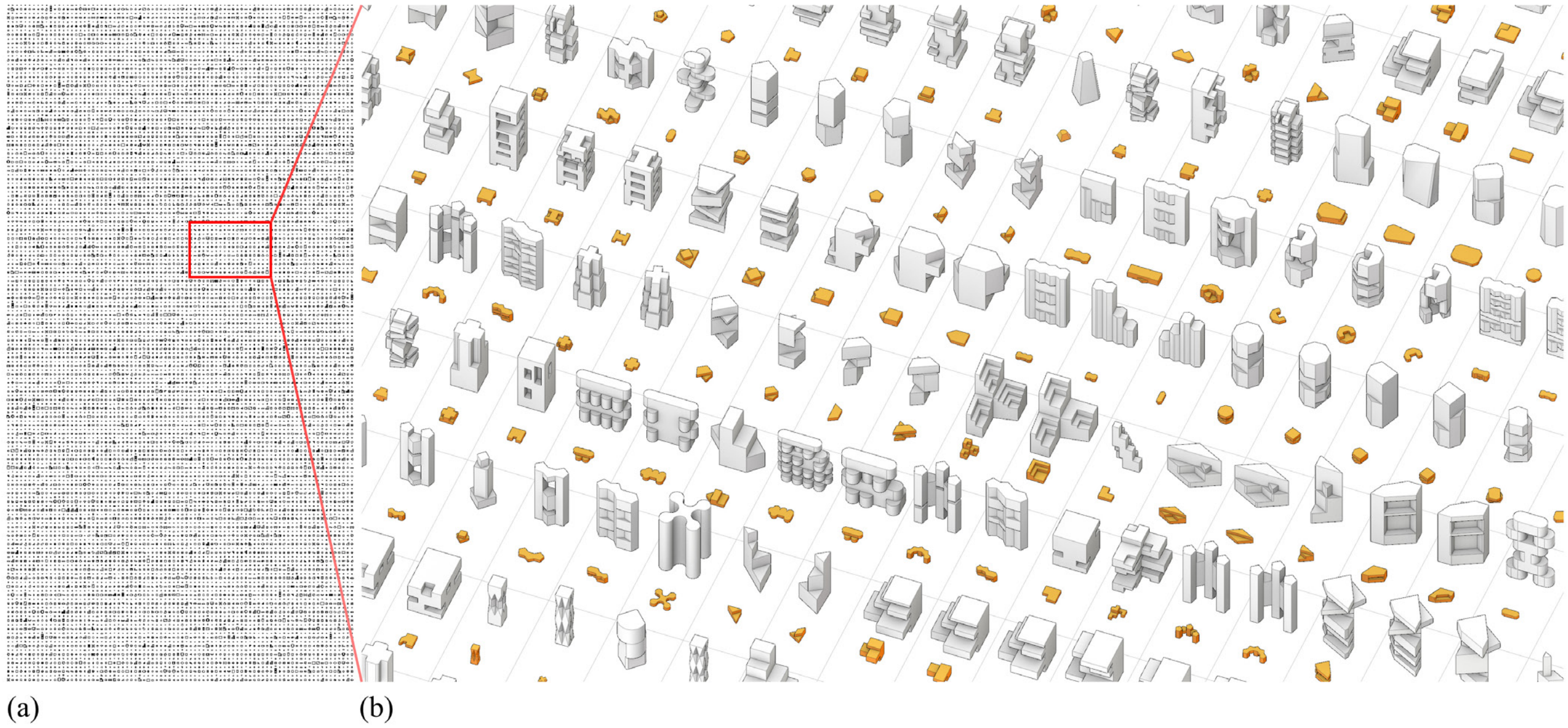

8

8 The full corpus at two scales. (a) Overview of all 6,505 prototype-massing pairs shown as perspective models. (b) Detail with massings in gray and prototypes in yellow.

A corpus of 1,301 base pairs was authored through typological study under a shared labeling protocol rather than collected from real projects. Prototype graphs contain 2-96 nodes and massing graphs 16-592 nodes, spanning simple prismatic envelopes through articulated forms.

Geometric perturbation produced 5,204 variants, for 6,505 graph pairs and 971,736 labeled nodes (Figure 8). Training used an unseeded random 85/15 record-level split that did not group authored and augmented relatives. The held-out set therefore measures within-corpus retrieval, not transfer to unseen design families.

### Baseline and Evaluation

We compare the learned model with direct nearest-neighbor retrieval after L2-normalizing each stored seven-dimensional node-feature vector. The HeteroGAT consumes the stored component-scaled inputs directly and L2-normalizes its learned 256-dimensional outputs. The reference uses no message passing and no learned transformation, so it tests the complete learned model against direct retrieval on the same descriptors. It does not bound what other geometry-only methods could achieve.

Accuracy is pooled over valid prototype anchors: each prototype node retrieves the nearest massing node, and correct predictions are divided by all 14,005 anchors from the 976 held-out pairs. Every prototype label occurs in its paired massing, so no anchor is excluded.

The best observed validation epoch correctly retrieved 13,514 of 14,005 anchors, or 96.5%. The reported raw-feature baseline reached 76.5% on the same held-out pairs, retrieval direction, and pooled aggregation. Because historical split membership and baseline logs did not survive, this remains a reported comparison rather than a re-executable one. The comparison is an existence test against direct retrieval on identical descriptors: it supports the claim that learned contextual information contributes beyond raw local-feature similarity, and does not attribute the margin to any individual component. Pattern assignments were stored only in Rhino, so the operational severity of label errors remains unmeasured. Figure 9 illustrates transfer within one related family rather than an independent generalization test.

## DEPLOYING CORRESPONDENCE: FROM LABELS TO GEOMETRY

### Pipeline Overview

The propagation stage converts retrieved labels into modeled facade geometry. Each target label indexes a prototype-side Pattern that stores module geometry, repetition, and placement directives. Correspondence determines which authored Pattern is routed to a segment; module design and geometric instantiation remain explicit downstream operations.

The complete pipeline is implemented in Rhino 8 (Robert McNeel & Associates 2026) as a demonstrator (Figure 10). A host extracts segment-level envelope curves, stores reusable module instances, and applies segment-relative placement transforms. Ingestion, inference, and propagation complete in seconds on the tested workstation, depending on geometry complexity. The four built-project massing models discussed below are qualitative applications rather than benchmarked evaluations.

### Authoring Phase

The designer constructs a representative facade fragment by arranging module geometries into one or more enclosed contour levels. No annotation schema, label assignment, or explicit propagation rule is involved; the spatial arrangement of placed modules is itself the specification.

From this arrangement, the system performs two concurrent parsing operations during prototype ingestion. First, it reconstructs the enclosed contours formed by the module layout and subdivides them into segments,

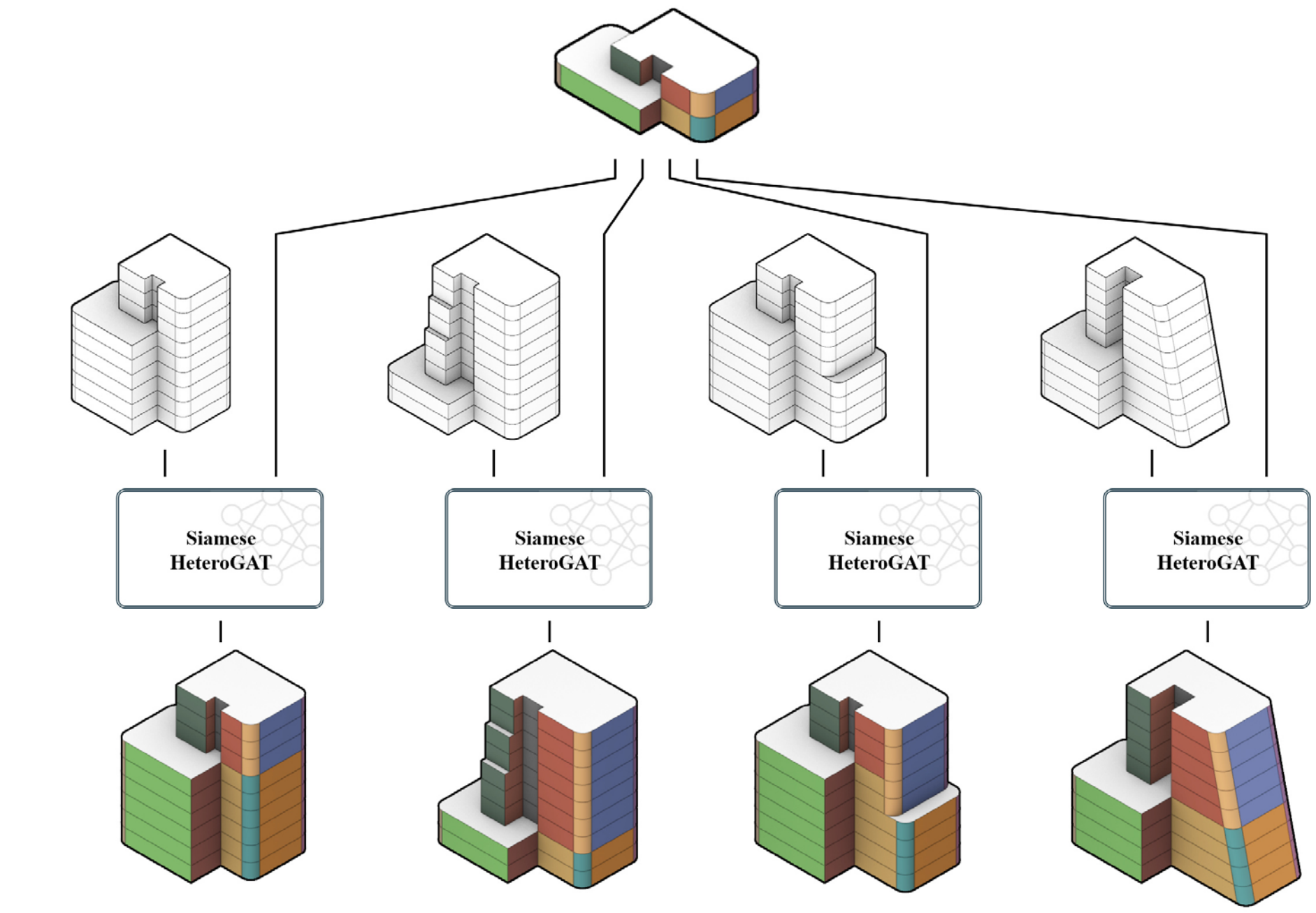

9

9 Illustrative family-internal articulation transfer. One prototype is paired with four related targets that share upper geometry and vary at the base. Predicted panel colors show pair-local correspondence; this example visualizes the workflow but is not an independent family-generalization test.

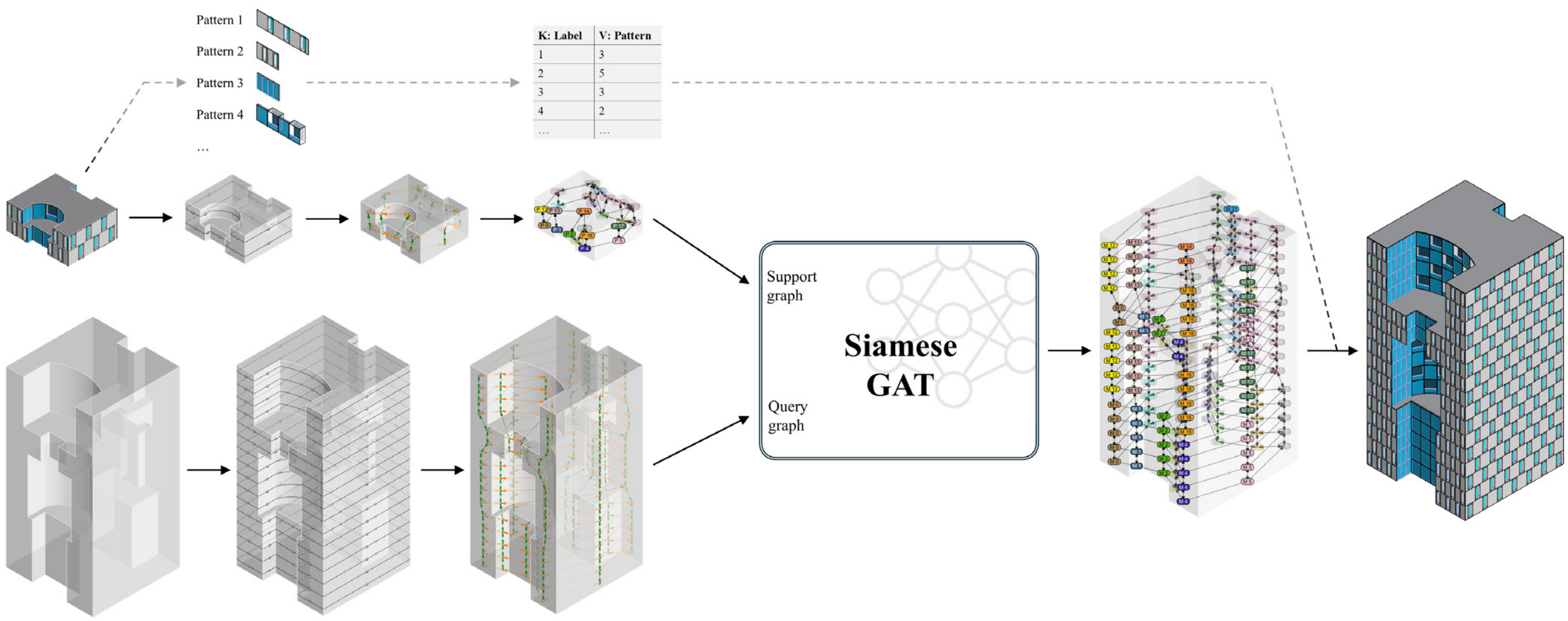

10

10 End-to-end deployment pipeline: Prototype ingestion (upper-left); massing ingestion (lower-left); routing-table construction (dashed); inference at the shared Siamese encoder (center); propagation onto the target massing (right).

constructing the support graph $G_s$ and its node set $V_s$. Second, it groups the module geometries by segment membership, extracting from each group the count, type, sequence, and spatial layout of modules it contains. This per-segment record constitutes the Pattern that the propagation stage will reproduce downstream.

A completed prototype must contain placed module geometry; without it, the system can extract contours but cannot recover Pattern definitions. Module-to-segment assignment is deterministic once a compatible exemplar has been constructed. The workflow avoids separate correspondence annotation and project-wide propagation-rule scripting, but still requires module and Pattern authoring, compatible segmentation, graph extraction, and host integration. Authoring time and usability were not evaluated.

### Pattern Instantiation and Propagation

Pattern instantiation reads the minimal repeating unit encoded in the Pattern: rather than a literal fixed-length module sequence, each Pattern stores the smallest subsequence whose repetition reproduces the prototype arrangement (A-B-A-B is encoded as the unit A-B), with the repetition count determined by the target segment's geometry at runtime. What transfers is relational structure, the minimal motif and its organizational logic, not a fixed geometric layout.

Length divergence is resolved by the Pattern's alignment directive, which anchors the module array to the segment (left, right, or centered) as the designer authored it. Residual length is absorbed by a designated atypical module at the alignment boundary, itself a Pattern-encoded preference. The target segment receives the prototype's spatial logic, not a geometric copy of its dimensions (Figure 11).

The computed module sequence and transforms then instantiate designer-authored module geometry onto the massing; propagation redistributes

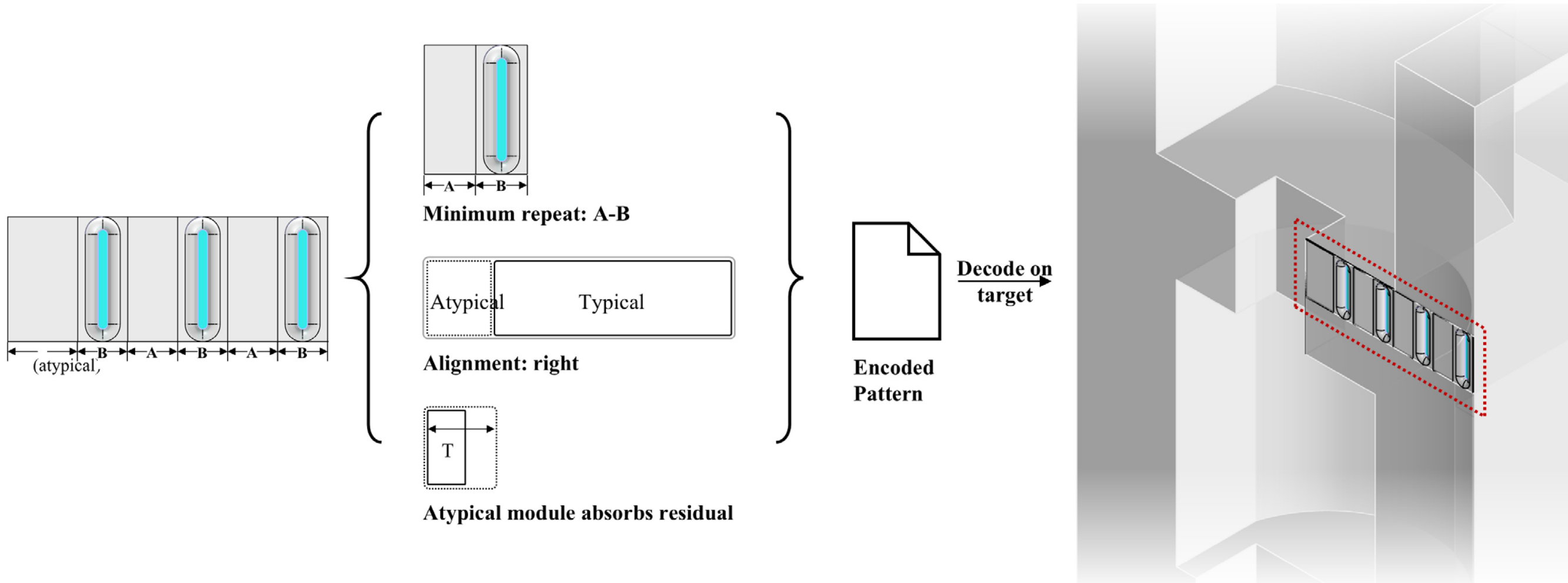

11

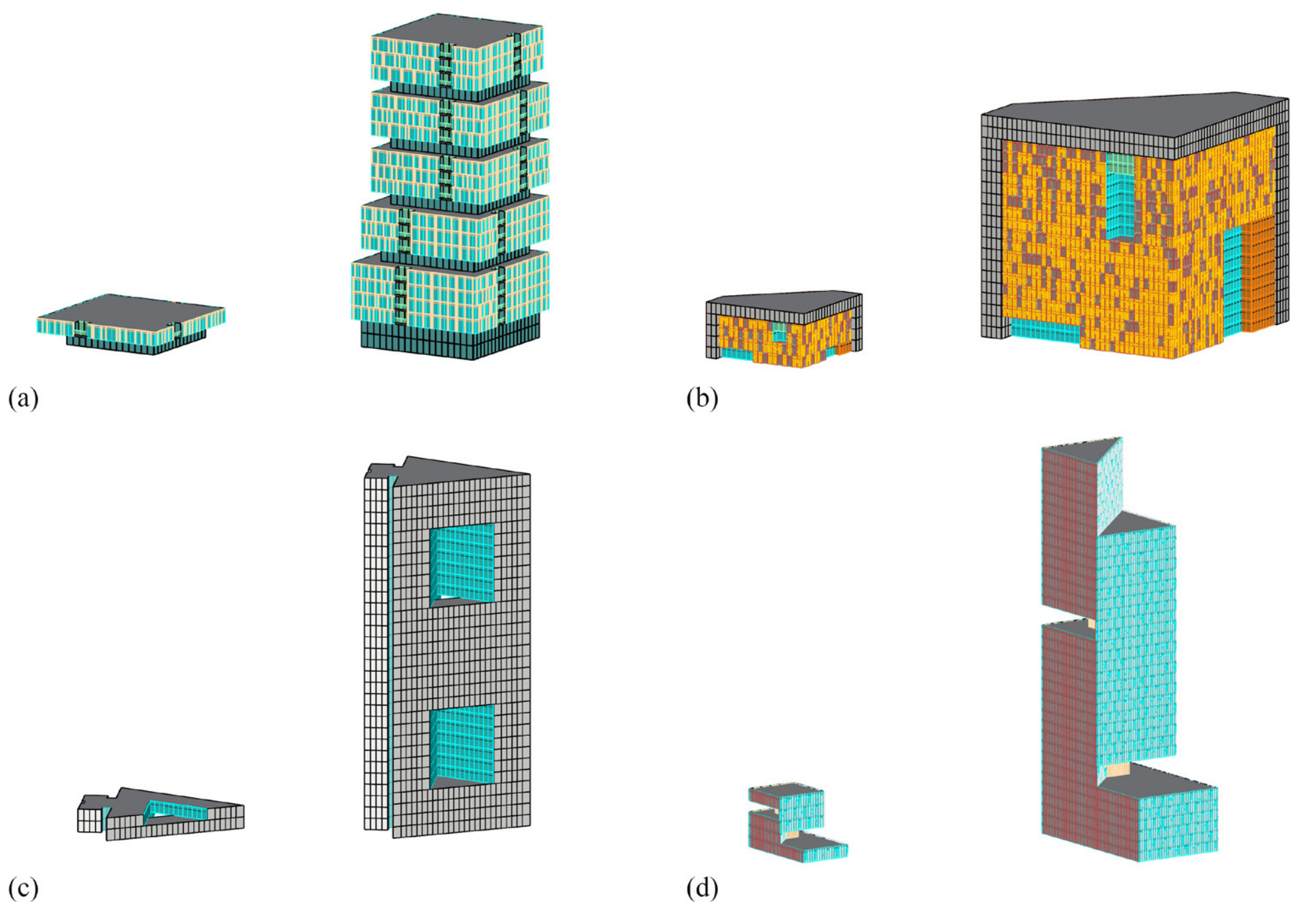

12

11 Pattern encoding and instantiation. A prototype segment's module sequence is decomposed into its minimal repeating unit, alignment directive, and atypical-module preference. The encoded Pattern is decoded onto a target segment, reproducing the sequence at the target's length with residual space absorbed at the alignment boundary.

12 Qualitative applications on massing models of four built projects: (a) One Steuart Lane, San Francisco; (b) Poly Corporation Headquarters, Beijing; (c) Manhattan Loft Gardens, London; and (d) National Commercial Bank, Jeddah. Each panel pairs a localized prototype with the propagated target. The cases show transfer across distinct massing organizations but are not benchmarked evaluations.

authored content rather than generating it.

### Applications to Built-Project Massings

The workflow was applied to prepared massing models of four built projects: One Steuart Lane, Poly Corporation Headquarters, Manhattan Loft Gardens, and National Commercial Bank (Figure 12). Each paired a localized prototype with a larger target under the same segmentation logic. The cases span setbacks, large openings, cantilevers, and recessed zones. They show Pattern routing across distinct massing organizations but are qualitative applications, not measurements of project-level accuracy or production reliability.

## DISCUSSION AND LIMITATIONS

### From Rule Authoring to Exemplar-Based Articulation

The framework addresses a bounded transfer problem: a designer has resolved a representative facade fragment, yet its intent has not been reconstructed as a project-wide rule system. It relocates the computational specification closer to where design intent already exists, in the authored geometry itself, while authorship remains with the designer.

The prototype defines a pair-local correspondence vocabulary; the shared encoder translates its segments across a compatible target, and downstream Patterns preserve explicit control over module content and placement. This division of labor distinguishes the workflow from both automatic facade generation and global semantic classification. One exemplar specifies deployment, but the encoder still derives its transferable metric from the multi-pair training corpus.

### Prototype Coverage as an Authoring Boundary

The prototype sets both the source and the limit of transfer (Figure 13). Withholding one facade condition from the exemplar, with the target and the trained encoder held fixed, ablates the specification

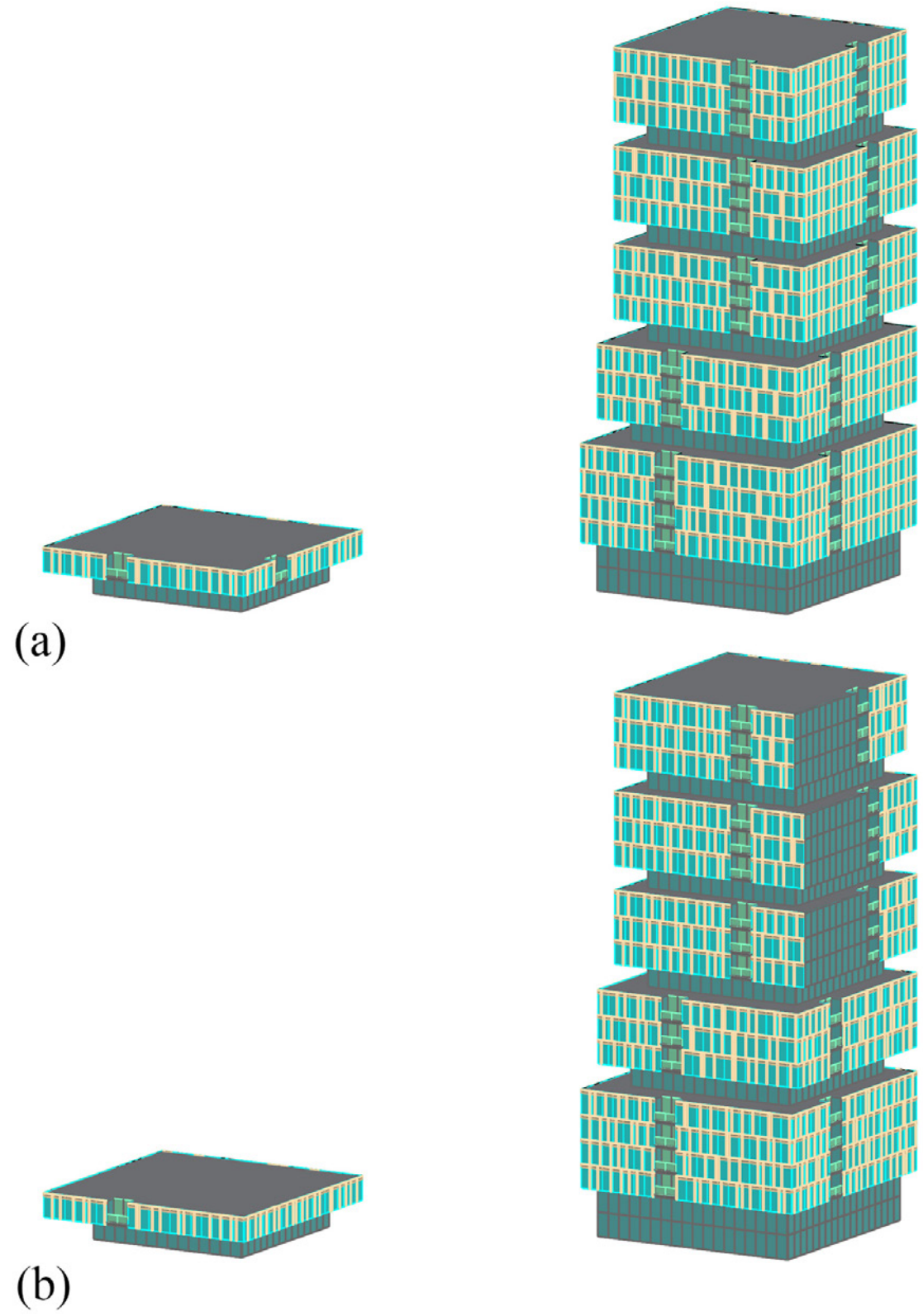

13

13 Effect of prototype coverage on One Steuart Lane. (a) A prototype representing the target's facade conditions supports complete Pattern transfer. (b) Omitting a facade condition leaves corresponding target regions under-resolved. Prototype coverage therefore defines the workflow's transferable design vocabulary.

rather than the model and isolates what the exemplar itself contributes. When the prototype represents the target conditions, Patterns propagate across the massing; when a condition is absent, affected segments receive the nearest available Pattern and remain under-resolved. Prototype coverage must therefore match the intended transfer scope. Training cannot recover content the exemplar never specifies.

Evidence remains bounded by a designer-authored synthetic corpus and a historical within-corpus holdout. The built-project cases use prepared massing models and are qualitative, not measured deployment trials. The reported accuracy measures the evaluation direction, in which prototype anchors retrieve from the massing; deployment reverses that query, and the paper demonstrates that direction rather than quantifying it. The evidence covers the full corpus range of prototype and massing sizes without resolving how accuracy varies with graph size inside it; size-dependent behavior is a distinct question from the authoring proposition advanced here. The study does not test survey or import noise, modeling tolerances, degenerate or non-manifold geometry, mixed segmentation logic, authoring usability, or production conditions. It establishes feasibility under the reported representation, not production reliability or generalization beyond the authored corpus.

## CONCLUSION

Facade modeling often falls between manual replication and project-specific rule authoring. This work demonstrates a third, bounded workflow: a designer resolves a local facade fragment, and a learned correspondence model routes its reusable Patterns across a compatible target envelope.

On the original within-corpus holdout, the complete model reached 96.5% prototype-to-massing accuracy, about 20 points above direct descriptor retrieval. The Rhino demonstrator and four built-project applications show how correspondence can route authored Patterns across distinct massing organizations, while the coverage comparison makes the principal boundary visible: the system cannot transfer facade intent absent from the exemplar.

The central contribution is an authoring proposition. Detailed facade geometry remains designed, modules and applicability boundaries remain explicit, and the system learns only how to align authored local intent with a larger relational structure. A geometric exemplar can therefore function as a computational specification without first being rewritten as a global generative rule.

## ACKNOWLEDGMENTS

This research was supported by SOM Research + Innovation and SOM IW. The author would like to extend particular thanks to Pablo Ruiz and Scott Duncan for the project that occasioned this research, and for the unreserved support that made both the work and this paper possible. Grateful acknowledgement is also due to An-Tai Lu and Wenxuan Xie for their invaluable assistance in generating and labeling massing models; their contributions were essential to the pace and progress of this research. Yao Lu and Yuanben Gao brought a keen eye to the writing, and the prose is clearer for it.

**Generative AI disclosure.** Claude Sonnet 4.6 (Anthropic 2026) assisted in refining manuscript wording in March 2026. Nano Banana Pro, the Gemini 3 Pro Image model (Google DeepMind 2026), applied a style transfer to one region of Figure 1 on July 27, 2026, working from the author's own geometry. No technical content, result, or analysis was generated by AI, and the author reviewed and verified all AI-assisted text.

## IMAGE CREDITS

**All drawings and images by the author. Figure 1 incorporates a generative style transfer applied to the author's own geometry; see Acknowledgments.**

**Yiliang Shao** is a Senior Computational Design Specialist at SOM Chicago, where he leads research and development projects at the intersection of academic research and professional practice. His research interest is machine learning for architectural design and automation in design workflows. He is first-named inventor on a pending US patent, and was selected into Metropolis Chicago Arch30, a nominated cohort of about 30 future architecture leaders in Greater Chicago. He holds a Master of Science in Design from the Weitzman School of Design, University of Pennsylvania, where he also worked as a Research Assistant in the Polyhedral Structures Laboratory.